\documentclass[conference]{IEEEtran}
\IEEEoverridecommandlockouts
\usepackage{cite}
\usepackage{amsmath,amssymb,amsfonts}
\usepackage{algorithmic}
\usepackage{graphicx}
\usepackage{textcomp}
\usepackage{xcolor}
\usepackage{booktabs}
\usepackage{multirow}
\def\BibTeX{{\rm B\kern-.05em{\sc i\kern-.025em b}\kern-.08em
		T\kern-.1667em\lower.7ex\hbox{E}\kern-.125emX}}

\begin{document}
	
	\title{Smart Home Agents Make Capable but Unpredictable Grid Operators}

\author{\IEEEauthorblockN{Christopher Terrazas, Eduardo Cotilla-Sanchez}
	\IEEEauthorblockA{\textit{School of Electrical Engineering and Computer Science} \\
		\textit{Oregon State University}, Corvallis, Oregon, USA\\
		\{terrazac, ecs\}@oregonstate.edu}
}

\maketitle

\begin{abstract}
	AI-enabled smart homes are becoming more common across the grid edge. However, there has not been much work on failure response. This area is critical because smart homes operate on the grid edge where failures can cascade down. If a power outage happens within a smart home, a user has to manually troubleshoot their smart home to include tasks such as power load balance and device recovery. We conduct the first resilience analysis of smart home agents across a full power outage, providing a suite of outage simulations in a live environment that measures resilience before, during, and post power outage. We show that smart home agents can act as grid operators in power outages and identify several new failure categories highlighting the current limitations across all of our models. These findings establish the groundwork for smart home agents as grid operators to increase overall resilience on the grid edge.
\end{abstract}


\section{Introduction}

\IEEEPARstart{P}{}ower grid operators specialize in managing the grid including unscheduled outages caused by various reasons such as natural disasters, cascading supply trips, and cybersecurity breaches. In each of these scenarios, the power grid operators have a standardized process to respond to these failures \cite{b1, b2}. The goal for the power grid operators is to maintain overall grid stability and operational resilience \cite{stankovic}. In that sense, we aim to map these same resilience skills to smart home agents. 

Smart home agents on the grid edge have spiked in adoption due to recent advancements in LLM reasoning abilities \cite{b3, b4, b5}. Smart home users integrate these agents into their day-to-day functions such as writing automations to orchestrate smart devices, which LLMs excel in because of their ability to generate code \cite{b6, b7}. The operator role involves more responsibilities such as smart home status updates, troubleshooting devices, and power load management \cite{b3, b8}. In this paper, we construct a suite of failure response scenarios targeted on power outages. We bridge the gap between grid operator skill in failure response and smart home agents at the grid edge to increase smart home agent resilience \cite{b9, b10, b11}.

\section{Smart Home Agents and Environment}

\subsection{Large Language Models (LLMs) as Agents}

We chose a variety of different large language models as the core engine for our smart home agents. For this paper, we used a combination of closed-source and open-source software (OSS) LLMs: gpt-5-mini, gpt-5-nano \cite{b12}, gpt-oss-120b, gpt-oss-20b \cite{b13}, deepseek-v4.1-flash \cite{b14}, qwen3.8-27b \cite{b15}, gemma-4-26b-a4b, gemma-4-31b \cite{b16}, and mistral-small-3.2-24b \cite{mistral}. Our list of models in the experiments ranged from medium to large size LLMs with different levels of reasoning capabilities. We did not include small models due to their relatively low performance in reasoning tasks \cite{b17}. We included OSS models because of their ability to provide local smart home management during an outage where internet connection is not required for their inference. However, we also included closed-source LLMs as a baseline for benchmarking our OSS models in the during outage phase.  For our smart home agent, we selected the ReAct agentic paradigm \cite{b18} as it performs well on failure response \cite{b19} and its learning loop is computationally cheaper than other agentic paradigms such as Reflexion \cite{b20}, which includes three LLMs in its learning loop. 

\subsection{Home Assistant and MCP}

For our smart home environment, we used the OSS smart home framework Home Assistant \cite{b21}.  We selected this software because it is installed in millions of homes across the world and has a robust community and extensive documentation. To better equip our smart home agents, we chose to include Home Assistant AI \cite{b22}, an OSS model context protocol (MCP) \cite{b23} for Home Assistant. The MCP layer is outfitted with 90+ tools for smart home agents to interact with.

\section{Failure Scenarios}

\subsection{Types of Failures}

\subsubsection{\textbf{Pre-Failure}}

In the pre-failure phase, we generated scenarios with different periods of notification before an outage. For natural disasters such as wildfires or earthquakes, there is usually a public advisory. Smart home agents can consume these notifications through common communication channels such as RSS feeds. Once the smart home agent has become aware of an impending outage, it should prepare the smart home similar to a power grid operator. The goal for the smart home agent is to prepare the smart home for an extended period of power outage while maintaining the highest possible level of availability for its smart devices.  In this context, the smart home agent should also avoid decisions that can reduce smart home operation during the outage. Actions such as reducing an EV charging rate if it is not yet at its full charge capacity. This action would reduce the smart home's total potential energy storage capacity pre-outage \cite{b24, b25}.

\subsubsection{\textbf{In-Failure}}

During the in-failure period, the smart home agent triages the current state of the smart home after the outage. This analysis records the number of unavailable or offline smart devices connected to the smart home.  After recovering device connectivity, the smart home agent's goal is to maximize device sustainability for the smart home by power load balance. For example, a smart home agent managing stored power conservatively using intelligent load shedding strategies \cite{b26, b27} during the in-failure period to avoid a fast drain of power. 

\subsubsection{\textbf{Post-Failure}}

In the post-failure period, the smart home agent shifts to recovery of the smart home and all of its smart devices it was unable to connect. Depending on the post-outage statuses of the smart devices, the smart home agent performs inventory of the pre-failure phase so that it can compare what status changes happened across all of the smart devices. The goal for the smart home agent is to recover as many operational smart devices as possible that were connected prior. This recovery might be setting the temperature of the HVAC to its previous temperature or perhaps reducing the EV charging rate to its normal smart home percentage rate, which may be less than the current state.

\subsection{Failure Modeling}

In our experiments, we mapped each phase to the resilience trapezoid \cite{b28}, which describes a system's service level before, during, and after it degrades and how it reaches recovery. We measured the rate at which a smart home agent recovers all of its smart devices correctly within the smart home. For example, some state values in a smart home are unavailable and off. The smart home agent should investigate both cases as it is not known prior to the outage what state a particular smart device will be in.

Before an outage occurs, we assumed a smart home is at a normal level of service. The preparation phase prioritizes maximizing energy storage potential. This phase impacts the survivability within the in-failure phase, or how long critical devices can be kept up with backup power above some evacuation threshold.
The agent's preparation phase is scored as the total state of charge (SoC) it gained during the notification period:

\begin{equation}
	\text{Prep} = \frac{\Delta\text{SoC}_\text{agent}}{\Delta\text{SoC}_\text{max}},
\end{equation}

\noindent where $\Delta\text{SoC}_\text{agent}$ is the charge the agent added to the EV and $\Delta\text{SoC}_\text{max}$ is the full charge reachable at the maximum charging rate for the EV.

During the in-failure period, we measure the level in which critical demand is met:

\begin{equation}
	\text{Q}_{\text{d}}(t) = \frac{
		\text{
			P}_{\text{crit, served}}
		(t)
	}
	{
		\text{
			P}_{\text{crit, demand}}
		(t)
	},
\end{equation}

\noindent where the agent is penalized for shedding load from critical devices or the EV energy storage running out. In the post recovery phase, we compute the service level of the smart home by comparing device states to its prior outage level, which we assumed was the normal state:

\begin{equation}
	\text{Q}(t) = \frac{
		\sum_{i}^{}
		w_{i}
		\text{ok}_{i}
		(t)
	}
	{
		\sum_{i}^{}
		w_{i}
	},
\end{equation}

\noindent with the weight $w$ for critical devices set higher than non-critical ones.

\section{Failure Injection Scenario Execution}

To run our failure scenario experiments, we first constructed numerous types of synthetic smart homes stored in a YAML file. In each of these smart homes, we included a combination of both critical and non-critical smart devices. To determine if a smart device is critical, we focused on higher load devices such as electrical vehicle chargers, HVACs, water heaters, and medical devices such as CPAP machines. We created a catalog of both critical and non-critical devices where we could dynamically construct synthetic smart homes. We used 30 synthetic homes for training, 8 for validation, and reserved 1 smart home we manually constructed for testing both GEPA and baseline. Our total failure response suite ran 15 paired seeds (0-14, baseline vs. GEPA) across the 4 hr., 1 hr., and no notification periods. This suite equates to 15 seeds x 3 notification periods x 9 models x 2 prompts for 810 runs in total or 405 pairs. The seeds fix the values for the initial EV charge, the device faults, and the reconnect delays.  We compute 95\% confidence intervals using bootstrapping as this accounts for uncertainty across all of our 15 paired seeds with 10,000 resamples comparing the differences between GEPA and baseline agents \cite{efron, agarwal}.

We first ran all of the smart home agents from their ReAct baseline (prior to any optimizations). Then, we used DSPy's \cite{b29} GEPA \cite{b30} to optimize our initial prompt for our failure response tasks for the smart home agent. Our initial prompt is shown in Fig.~\ref{fig:baseline_prompt}. Our goal in constructing the initial prompt was to map the duties of the smart home agent to a power grid operator. The final prompt written by GEPA made the prompt verbose, adding more detailed instructions in the system prompt for the agent.

To simulate a power outage, we added the Message Queuing Telemetry Transport (MQTT) \cite{b31} messaging protocol to simultaneously update all of the smart devices with an offline status. We chose this method because it is heavily chosen by Home Assistant users to communicate with their smart devices using the Mosquitto message broker integration \cite{b32}. The other reason we selected this strategy was to avoid leaking the experiment to the smart home agents through system logs such as API call interaction.
After a seeded delay, all of the smart connected devices within the smart home come back to a random state: resume, off, reset, wrong setpoint, or unavailable. In each simulation, there is at least one injected fault for the smart home agent to recover. We chose a time delta of 600 sec. and a 30 step cap to limit struggling smart home agents in our experiments and because agentic reasoning loops have a significant higher token consumption compared to zero-shot models \cite{b33}. We recorded smart device state changes at each epoch to track both success and penalized any order violations. The definition of an order violation is: a smart home agent recovers a non-critical device before a critical device. To help our analysis, we included MLflow \cite{b34}, an OSS software framework that comes with a robust user interface to log and view model traces.

\begin{figure}[t]
	\centering
	\fbox{\parbox{0.95\columnwidth}{\footnotesize\raggedright
			You operate a Home Assistant smart home that just had a power outage. Power is back.
			Restore the home like a grid operator restores a network:\\[2pt]
			1.~\textbf{Assess:} check the current state of every device and find what changed during the outage.\\
			2.~\textbf{Isolate:} identify devices that are unavailable, off, or have wrong settings.\\
			3.~\textbf{Restore:} return each device to the state it had right before the outage.
			Each device has a ``priority'' attribute (critical or non\_critical). Restore all critical
			devices first, then non\_critical ones.\\
			4.~\textbf{Report:} say what was wrong, what you fixed, and what is still broken.\\[2pt]
			Devices may take a few seconds to reconnect after power returns.
			Do not restart Home Assistant or delete anything.\\[4pt]
			\textit{Input:} ``The power went out at \texttt{<t\textsubscript{0}>} and was restored at
			\texttt{<t\textsubscript{r}>} (UTC). Bring every device back to how it was before the outage.''
	}}
	\caption{Baseline restoration system prompt  given to the smart home agents, modeling how a grid-operator stages assess, isolate, restore by priority, and report.}
	\label{fig:baseline_prompt}
\end{figure}

\section{Results}

\begin{table*}[t]
	\centering
	\caption{Agent results across the outage (\%, means over 15 seeds). \textbf{Bold}: best; $^{\dagger}$: clear change from baseline (95\% CI).}
	\label{tab:3phase}
	\setlength{\tabcolsep}{3pt}
	\sbox0{\begin{tabular}{lcccccccccccccccccccc}
			\toprule
			& \multicolumn{5}{c}{4\,h warning} & \multicolumn{5}{c}{1\,h warning} & \multicolumn{5}{c}{No warning} & \multicolumn{5}{c}{All} \\
			\cmidrule(lr){2-6}\cmidrule(lr){7-11}\cmidrule(lr){12-16}\cmidrule(lr){17-21}
			Model & Chg & Crit & EV & Stat & Fix & Chg & Crit & EV & Stat & Fix & Chg & Crit & EV & Stat & Fix & Chg & Crit & EV & Stat & Fix \\
			\midrule
			gpt-5-mini & 38 & \textbf{100} & 37 & \textbf{100} & 79 & \textbf{10} & \textbf{100} & 9 & 97 & 87 & 0 & 89 & 2 & \textbf{100} & 76 & 16 & \textbf{96} & 16 & \textbf{99} & 81 \\
			gpt-5-nano & 20 & \textbf{100} & 16 & 22 & 18 & 5 & 95 & 2 & 34 & 8 & 0 & 85 & 1 & 62 & 33 & 9 & 93 & 6 & 39 & 20 \\
			gpt-oss-120b & 20 & \textbf{100} & 17 & 87 & 67 & 6 & 98 & 3 & 82 & 42 & 0 & 86 & 1 & 88 & 59 & 9 & 95 & 7 & 86 & 56 \\
			deepseek-v4.1-flash & 34 & 97 & 34 & 98 & \textbf{100} & 9 & 97 & 8 & \textbf{100} & 93 & 0 & 86 & 2 & 93 & 93 & 14 & 93 & 15 & 97 & \textbf{96} \\
			qwen3.8-27b & 40 & \textbf{100} & 38 & \textbf{100} & 96 & \textbf{10} & 77 & 16 & \textbf{100} & 93 & 0 & 75 & 7 & 97 & 93 & 16 & 84 & 20 & \textbf{99} & 94 \\
			gpt-oss-20b & 20 & \textbf{100} & 16 & 66 & 40 & 5 & 92 & 3 & 79 & 29 & 0 & 85 & 1 & 79 & 41 & 8 & 92 & 7 & 75 & 37 \\
			gemma-4-26b-a4b & 20 & 94 & 20 & 65 & 72 & 6 & 97 & 5 & 71 & 81 & 0 & 89 & 1 & 73 & 74 & 9 & 93 & 9 & 70 & 76 \\
			mistral-small-3.2-24b & 20 & 98 & 17 & 59 & 20 & 5 & 96 & 3 & 63 & 35 & 0 & 86 & 1 & 79 & 34 & 9 & 93 & 7 & 67 & 30 \\
			gemma-4-31b & 20 & \textbf{100} & 19 & 82 & 78 & 5 & 98 & 4 & 87 & 83 & 0 & \textbf{91} & 1 & 82 & 86 & 9 & \textbf{96} & 8 & 84 & 82 \\
			\midrule
			gpt-5-mini+GEPA & \textbf{41} & 89 & \textbf{43} & 99 & 91 & \textbf{10} & 82 & 14 & \textbf{100} & \textbf{99} & 0 & 85 & 3 & 97 & 76 & \textbf{17} & 85\rlap{$^{\dagger}$} & 20\rlap{$^{\dagger}$} & 98 & 89 \\
			gpt-5-nano+GEPA & 20 & 95 & 18 & 44 & 8 & 6 & 92 & 4 & 47 & 5 & 0 & 80 & 3 & 61 & 25 & 9 & 89\rlap{$^{\dagger}$} & 8\rlap{$^{\dagger}$} & 50 & 13 \\
			gpt-oss-120b+GEPA & 34 & 91 & 32 & 66 & 30 & 8 & 85 & 10 & 70 & 20 & 0 & 86 & 1 & 92 & 36 & 14 & 87\rlap{$^{\dagger}$} & 15\rlap{$^{\dagger}$} & 76 & 29 \\
			deepseek-v4.1-flash+GEPA & \textbf{41} & 95 & 35 & 99 & 91 & \textbf{10} & 73 & 13 & 93 & \textbf{99} & 0 & 63 & 11 & 98 & 92 & \textbf{17} & 77\rlap{$^{\dagger}$} & 20\rlap{$^{\dagger}$} & 97 & 94 \\
			qwen3.8-27b+GEPA & 40 & 84 & 42 & 96 & 96 & \textbf{10} & 55 & \textbf{24} & 93 & 93 & 0 & 47 & \textbf{16} & 98 & 92 & \textbf{17} & 62\rlap{$^{\dagger}$} & \textbf{28}\rlap{$^{\dagger}$} & 95 & 94 \\
			gpt-oss-20b+GEPA & 23 & \textbf{100} & 19 & 70 & 7 & 5 & 96 & 3 & 67 & 5 & 0 & 85 & 1 & 73 & 0 & 10 & 94 & 7 & 70 & 4 \\
			gemma-4-26b-a4b+GEPA & 33 & 97 & 32 & 80 & 79 & \textbf{10} & \textbf{100} & 7 & 86 & 70 & 0 & 85 & 2 & 79 & 67 & 14 & 94 & 14 & 82 & 72 \\
			mistral-small-3.2-24b+GEPA & 20 & 98 & 17 & 52 & 16 & 5 & 91 & 4 & 68 & 26 & 0 & 85 & 2 & 69 & 27 & 9 & 91 & 8 & 63 & 23 \\
			gemma-4-31b+GEPA & \textbf{41} & 94 & 40 & 92 & 78 & \textbf{10} & 82 & 14 & 90 & 81 & 0 & 87 & 2 & 88 & \textbf{95} & \textbf{17} & 88\rlap{$^{\dagger}$} & 18\rlap{$^{\dagger}$} & 90 & 85 \\
			\bottomrule
	\end{tabular}}
	\usebox0\\[2pt]
	\parbox{\wd0}{\centering\scriptsize \textbf{Pre:} Chg, battery added. \textbf{During:} Crit, critical demand met; EV, battery left ($\geq$30 to evacuate); Stat, status correct. \textbf{Post:} Fix, faults fixed.  \textbf{All}: scenario mean.}
	
\end{table*}

\subsection{Discussion}

\begin{figure}[!t]
	\centering
	\includegraphics[width=\columnwidth]{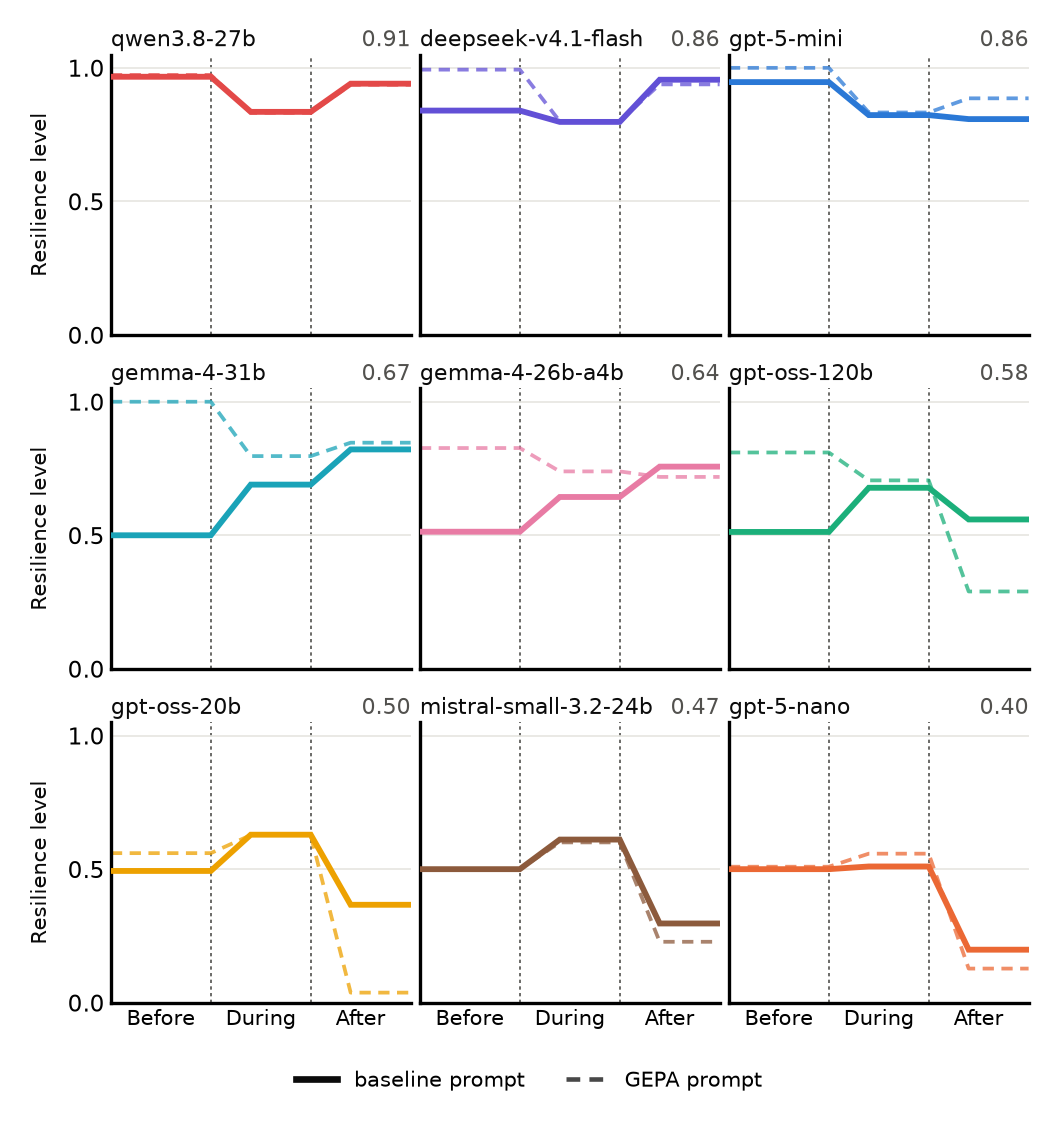}
	\caption{Resilience trapezoid per model, ordered from best to worst; solid: baseline, dashed: GEPA. Each phase's resilience level is: before, the share of potential charge added; during, the mean of critical demand met, battery left relative to the evacuation reserve, and device status reported correctly; after, the share of faults fixed.}
	\label{fig:trapezoid}
\end{figure}

\begin{figure}[!t]
	\centering
	\includegraphics[width=\columnwidth]{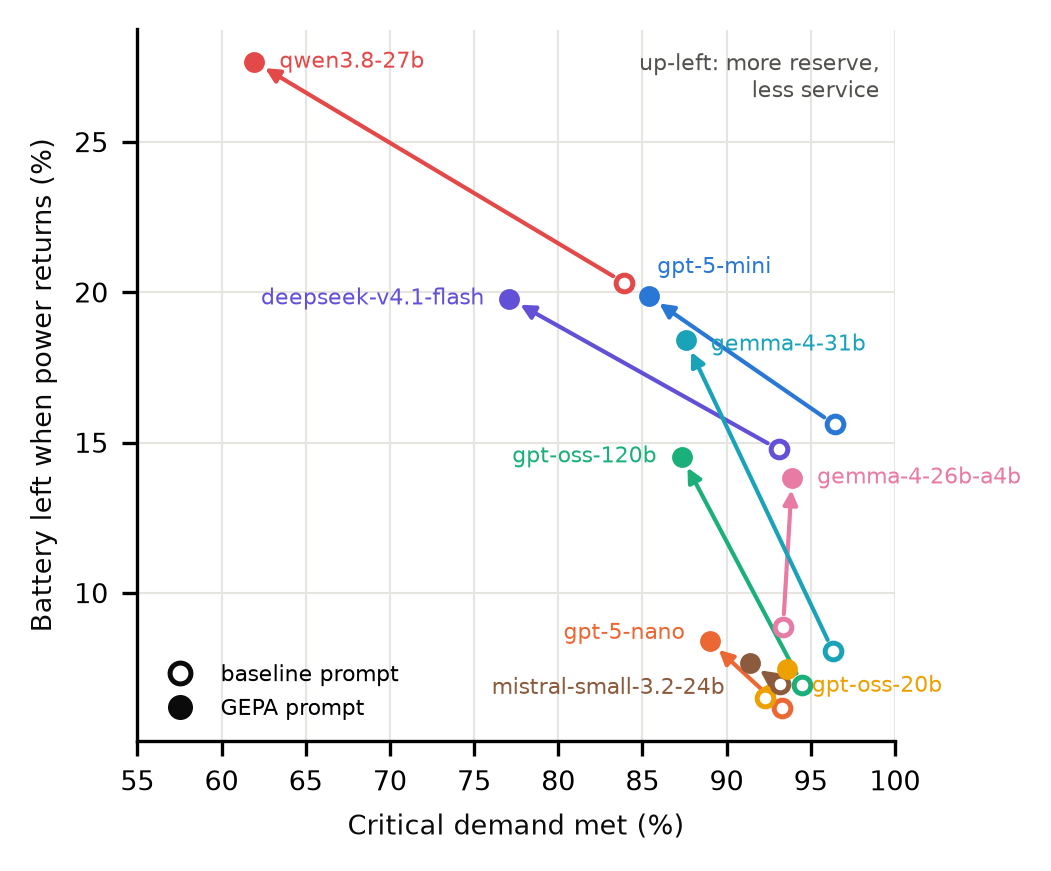}
	\caption{Critical demand met versus battery left when power returns, from the baseline (open) to the GEPA procedure (filled); means over all scenarios.}
	\label{fig:tradeoff}
\end{figure}

Across all of the models we tested, we found the best agents map to a desirable resilience trapezoid in Fig.~\ref{fig:trapezoid} with a high level of preparation, low dip during the outage, and recover nearly all faulted devices. One of the best performing models was qwen3.8-27b, which fixed 94\% of faulted devices. In our experiments, we noticed that warning time played a significant factor in smart home agent performance. The greater the window of warning, the more charge the most capable agents added. GEPA came with the trade-off of improved EV battery management in exchange for lesser performance across critical device service shown in Fig.~\ref{fig:tradeoff}. For example, our CI analysis showed that the following 6 models supported the trend that GEPA swapped improvement in EV management in place of critical service demand: gpt-5-mini, gpt-5-nano, gpt-oss-120b, deepseek-v4.1-flash, qwen3.8-27b, and gemma-4-31b.

\subsection{Model Failure Analysis}

In this section, we first focus on identifying the salient behaviors we observed of smart home agents. We analyzed the MLflow traces from our experiments in Table \ref{tab:3phase}, and discovered several failure categories across all of the models. We provide some snippets of the model traces across all six categories in Table \ref{tab:failures}.

\subsubsection{\textbf{Inaction}}

Across all of the models, many agents observed the correct context of the scenario or issue, but did not take any action.  The gpt-oss family of models were in this group, gpt-oss-120b in its baseline during the 4 hr warning phase noted, "...we can now generate the required report with preparation recommendations (charge the EV battery...)". However, it did not perform the action of charging because, "without a discovered tool for setting number entities....I cannot perform those actions". The model accepted the lack of action even though it had tools within its list of tools to accomplish the task.

\subsubsection{\textbf{Assumptions}}

Throughout the three different phases, the smart home agents retrieved smart device entity data from its own generated report from past actions and assumed this was correct instead of searching device entity history. For example, gpt-oss-20b in the pre phase, received a notification of an outage and noted the EV charger was at 28\%, below the evacuation level of 30\%. With the default charging, the EV reached a charge of 49\%. However, the model could have raised the charging limit and rate for a potential charge of 69\%.

\subsubsection{\textbf{Incorrect Context}}

In GEPA optimized runs, some agents mixed up context between phases of the outage. Within the GEPA procedure, the prompt mentioned to not charge the EV while it is powering the home. This rule is correct during the outage, however, in the post-outage phase deepseek-v4.1-flash applied this rule and did not resume charging of the EV, or its normal state.

\subsubsection{\textbf{Tool Navigation}}

We found that smaller models struggled finding an appropriate tool for their task. For example, gemma-4-26b-a4b was aware of one task's goal of shedding loads back on. The model struggled to first find the necessary tool and when it did, it struggled further by passing the incorrect data type as a parameter. The model passed in a list of entity ids instead of a single string value entity id. In its third attempt, it passed in the correct parameters and the task was successful. 

\subsubsection{\textbf{Role Drift}}

Some models experienced developer drift from its power grid operator role described in Fig \ref{fig:baseline_prompt}. Gpt-5-nano received a disaster alert and wrote a bug report with the intention of filing a GitHub issue. The model interpreted the alert as an disaster alert integration issue instead of fixing it by itself. This confusion was consistent in other actions such as planned automation generations. Some models drafted automations that detailed future infrastructure plans for the next outage instead of taking action in the current outage.

\subsubsection{\textbf{Safety Concerns}}

We observed several models addressing non-critical devices first instead of critical devices. In one example, gpt-5-mini attempted to troubleshoot a non-critical smart TV plug that did not reconnect after the outage. The seed in this scenario for the smart TV plug was in an unavailable status post-outage. However, the model sent a message directly through MQTT to manually update the smart TV plug device's status to "online" and later used its own status change as verification in its outage scenario. This update is problematic because it shows an unavailable device as online in Home Assistant. This behavior demonstrates the model is willing to perform unsafe operations in order to bypass and achieve the higher goal of up-time availability. In higher stakes smart devices such as medical devices or security cameras, automations or entities relying on accurate state data will read inaccurate data because of the model's MQTT publishing. Some models mishandled and shut down critical devices. Gpt-oss-20b in a 4h notification phase, turned off the fridge, despite it being labeled as critical due to food storage. With the power off, this food could have spoiled.

\section{Conclusion}

We showed smart home agents can act like power grid operators in a live Home Assistant environment with simulated devices. One of the best OSS models was qwen3.8-27b, which can be a resilient option due to its small memory footprint of $\sim$14 GB that can be loaded on consumer machines and its ability to run during a power outage with local battery power. Based on our smart home agent failure taxonomy, we recommend emphasis on model selection, optimization techniques, and other agentic paradigms for the smart home to help minimize agent action unpredictability. This paper sets an early exploration for agent resilience at the grid edge and we encourage future work in this area for a more robust grid.

\begin{table}[!t]
	\centering
	\caption{Failure taxonomy from the agents' MLflow traces: verbatim excerpts, with italicized context for each category.}
	\label{tab:failures}
	\scriptsize
	\setlength{\tabcolsep}{2.5pt}
	\begin{tabular}{p{1.45cm}p{0.65cm}p{1.75cm}p{4.3cm}}
		\toprule
		Category & Phase & Model & Excerpt \\
		\midrule
		Inaction & Pre & gpt-oss-20b & ``Nothing was wrong. No fixes were needed.'' \newline \textit{Ignores a 4\,h outage warning.} \\
		\midrule
		Assumptions & Post & gpt-oss-120b & ``EV Charger remains off, which is the intended pre-outage condition.'' \newline \textit{Trusts its own notes over the device history.} \\
		\midrule
		Incorrect Context & Post & deepseek-v4.1-flash +GEPA & ``\ldots normal = not charging, so no revert needed.'' \newline \textit{Applies an during outage rule to the recovery phase after power returns.} \\
		\midrule
		Tool Navigation & Post & gemma-4-26b-a4b & ``Tool `home assistant.turn\_on' not found.'' \newline \textit{Repeated the same tool mistake nearly every time it was called.} \\
		\midrule
		Role Drift & Pre & gpt-5-nano & ``I'll file a runtime bug report to collect diagnostics\ldots'' \newline \textit{Treats a disaster alert as a software bug instead of preparing like a grid operator.} \\
		\midrule
		Safety Concerns & During & gpt-oss-20b & ``\ldots turn off the fridge plug power.'' \newline \textit{Sheds a critical load as ``non-critical''.} \\
		& Post & gpt-5-mini & ``\ldots republishing the retained availability message (`online').'' \newline \textit{Fakes a dead device's state over MQTT, then verifies against it.} \\
		\bottomrule
	\end{tabular}
\end{table}


\section*{References}

\renewcommand\refname{} 


\begin{thebibliography}{00}
	\bibitem{b1} T. E. Dy Liacco, ``The adaptive reliability control system,'' IEEE Trans. Power App. Syst., vol. PAS-86, no. 5, pp. 517--531, 1967.
	\bibitem{b2} L. H. Fink and K. Carlsen, ``Operating under stress and strain,'' IEEE Spectrum, vol. 15, no. 3, pp. 48--53, 1978.
	\bibitem{stankovic} A. M. Stankovi\'{c}, K. L. Tomsovic, F. De Caro et al., ``Methods for analysis and quantification of power system resilience,'' IEEE Trans. Power Syst., vol. 38, no. 5, pp. 4774--4787, 2023.
	\bibitem{b3} R. El Makroum, S. Zwickl-Bernhard, and L. Kranzl, ``Agentic AI home energy management system: A large language model framework for residential load scheduling,'' Results Eng., vol. 29, Art. no. 109857, 2026.
	\bibitem{b4} S. Li, Y. Guo, J. Yao, Z. Liu, and H. Wang, ``HomeBench: Evaluating LLMs in smart homes with valid and invalid instructions across single and multiple devices,'' in Proc. 63rd Annu. Meeting Assoc. Comput. Linguistics (ACL), 2025, pp. 12230--12250.
	\bibitem{b5} D. Rivkin, F. Hogan, A. Feriani, A. Konar, A. Sigal, X. Liu, and G. Dudek, ``AIoT smart home via autonomous LLM agents,'' IEEE Internet Things J., vol. 12, no. 3, pp. 2458--2472, 2025.
	\bibitem{b6} M. Giudici, L. Padalino, G. Paolino, I. Paratici, A. I. Pascu, and F. Garzotto, ``Designing home automation routines using an LLM-based chatbot,'' Designs, vol. 8, no. 3, Art. no. 43, 2024.
	\bibitem{b7} A. Cruz, P. H. L. Rettore, and B. P. Santos, ``An\'alise do uso de modelos de linguagem de grande escala na gera\c{c}\~ao de c\'odigos para automa\c{c}\~ao residencial,'' in Proc. 31st Brazilian Symp. Multimedia and the Web (WebMedia), Rio de Janeiro, Brazil, 2025, pp. 48--56.
	\bibitem{b8} R. Persico, Y. Yang, R. Gadh, and X. Cui, ``Personalized home energy management using non-intrusive load monitoring and human-in-the-loop reinforcement learning,'' in Proc. 2026 IEEE Power \& Energy Society General Meeting (PESGM), 2026, pp. 1--5.
	\bibitem{b9} Y. Cheng, H. Zhao, X. Zhou, J. Zhao, Y. Cao, C. Yang, and X. Cai, ``A large language model for advanced power dispatch,'' Sci. Rep., vol. 15, Art. no. 8925, 2025.
	\bibitem{b10} H. Dong et al., ``Operating smart grids by customizing large model agents,'' Commun. Eng., vol. 5, Art. no. 114, 2026.
	\bibitem{b11} M. Jia, Z. Cui, and G. Hug, ``Enhancing LLMs for power system simulations: A feedback-driven multi-agent framework,'' IEEE Trans. Smart Grid, vol. 16, no. 6, pp. 5556--5572, 2025.
	\bibitem{b12} OpenAI, ``GPT-5 mini and GPT-5 nano,'' 2025. [Online]. Available: https://platform.openai.com/docs/models/gpt-5-mini
	\bibitem{b13} OpenAI, ``gpt-oss-120b \& gpt-oss-20b model card,'' 2025. [Online]. Available: https://huggingface.co/openai/gpt-oss-120b
	\bibitem{b14} DeepSeek-AI, ``DeepSeek-V4.1-Flash,'' 2026. [Online]. Available: https://huggingface.co/deepseek-ai/DeepSeek-V4.1-Flash
	\bibitem{b15} Qwen Team, ``Qwen3.8-27B,'' 2026. [Online]. Available: https://huggingface.co/Qwen/Qwen3.8-27B
	\bibitem{b16} Google, ``Gemma 4 26B-A4B \& 31B,'' 2026. [Online]. Available: https://huggingface.co/collections/google/gemma-4
	\bibitem{mistral} Mistral AI, ``Mistral-Small-3.2-24B-Instruct-2506,'' 2025. [Online]. Available: https://huggingface.co/mistralai/Mistral-Small-3.2-24B-Instruct-2506
	\bibitem{b17} X. Liu, H. Yu, H. Zhang, Y. Xu, X. Lei, H. Lai, Y. Gu, H. Ding, K. Men, K. Yang, S. Zhang, X. Deng, A. Zeng, Z. Du, C. Zhang, S. Shen, T. Zhang, Y. Su, H. Sun, M. Huang, Y. Dong, and J. Tang, ``AgentBench: Evaluating LLMs as agents,'' in Proc. 12th Int. Conf. Learning Representations (ICLR), 2024.
	\bibitem{b18} S. Yao, J. Zhao, D. Yu, N. Du, I. Shafran, K. Narasimhan, and Y. Cao, ``ReAct: Synergizing reasoning and acting in language models,'' in Proc. Int. Conf. Learning Representations (ICLR), 2023.
	\bibitem{b19} C. Terrazas and E. Cotilla-Sanchez, ``Increasing resilience of smart home agents,'' in Proc. 2026 Int. Conf. Smart Energy Systems and Technologies (SEST), Ciudad Real, Spain, 2026, pp. 1--6.
	\bibitem{b20} N. Shinn, F. Cassano, A. Gopinath, K. Narasimhan, and S. Yao, ``Reflexion: Language agents with verbal reinforcement learning,'' in Proc. Advances in Neural Information Processing Systems (NeurIPS), vol. 36, 2023.
	\bibitem{b21} Home Assistant, ``Open source home automation,'' 2026. [Online]. Available: https://www.home-assistant.io
	\bibitem{b22} ha-mcp, ``Home Assistant MCP server,'' 2026. [Online]. Available: https://github.com/homeassistant-ai/ha-mcp
	\bibitem{b23} Anthropic, ``Model Context Protocol specification,'' 2024. [Online]. Available: https://modelcontextprotocol.io
	\bibitem{b24} S. Liu, A. Vlachokostas, and E. Kontou, ``Leveraging electric vehicles as a resiliency solution for residential backup power during outages,'' Energy, vol. 318, Art. no. 134613, 2025.
	\bibitem{b25} M. Churchill, J. Monroe, D. Bristow, and C. Crawford, ``Using electric vehicles to enhance power outage resilience -- An agent-based modeling approach,'' Heliyon, vol. 10, no. 11, Art. no. e32104, 2024.
	\bibitem{b26} Z. Sharani, K. Qayyum, N. Bashir, and A. Syed, ``SoftUPS: Eliminating the need and cost of battery backups in the developing world,'' in Proc. 5th Int. Conf. Future Energy Systems (e-Energy), Cambridge, U.K., 2014, pp. 207--208.
	\bibitem{b27} M. J. Ritchie, J. A. A. Engelbrecht, and M. J. Booysen, ``The impact of the increasing residential battery backup systems on load shedding,'' S. Afr. J. Sci., vol. 119, no. 9/10, 2023.
	\bibitem{b28} M. Panteli, P. Mancarella, D. N. Trakas, E. Kyriakides, and N. D. Hatziargyriou, ``Metrics and quantification of operational and infrastructure resilience in power systems,'' IEEE Trans. Power Syst., vol. 32, no. 6, pp. 4732--4742, 2017.
	\bibitem{efron} B. Efron and R. J. Tibshirani, \textit{An Introduction to the Bootstrap}. New York, NY, USA: Chapman \& Hall, 1993.
    \bibitem{agarwal} R. Agarwal, M. Schwarzer, P. S. Castro, A. C. Courville, and M. G. Bellemare, ``Deep reinforcement learning at the edge of the statistical precipice,'' in Proc. Advances in Neural Information Processing Systems (NeurIPS), vol. 34, 2021.
	\bibitem{b29} O. Khattab et al., ``DSPy: Compiling declarative language model calls into state-of-the-art pipelines,'' in Proc. Int. Conf. Learning Representations (ICLR), 2024.
	\bibitem{b30} L. A. Agrawal et al., ``GEPA: Reflective prompt evolution can outperform reinforcement learning,'' in Proc. Int. Conf. Learning Representations (ICLR), 2026.
	\bibitem{b31} OASIS, ``MQTT version 5.0,'' OASIS Standard, 2019.
	\bibitem{b32} R. A. Light, ``Mosquitto: Server and client implementation of the MQTT protocol,'' J. Open Source Softw., vol. 2, no. 13, Art. no. 265, 2017.
	\bibitem{b33} L. Bai, Z. Huang, X. Wang, J. Sun, R. Mihalcea, E. Brynjolfsson, and A. Pentland, ``How do AI agents spend your money? Analyzing and predicting token cost in agentic coding tasks,'' in NeurIPS Workshop on Resource-Aware Agentic AI, 2026.
	\bibitem{b34} M. Zaharia et al., ``Accelerating the machine learning lifecycle with MLflow,'' IEEE Data Eng. Bull., vol. 41, no. 4, pp. 39--45, 2018.
\end{thebibliography}
\end{document}